\documentclass[runningheads]{llncs}
\usepackage{graphicx}

\usepackage{tikz}
\usepackage{booktabs}
\usepackage{comment}
\usepackage{amsmath,amssymb} 
\usepackage{wrapfig}
\usepackage{color}
\usepackage{url}
\usepackage{hyperref}
\hypersetup{
    colorlinks=true,
    linkcolor=red,
    filecolor=red,      
    urlcolor=red,
    }

\usepackage[accsupp]{axessibility}  

\begin{document}
\pagestyle{headings}
\mainmatter
\def\ECCVSubNumber{5901}  

\title{Object Wake-up: 3D Object Rigging from a Single Image} 



%
\author{Ji Yang\inst{1*} \and Xinxin Zuo\inst{1*} \and
Sen Wang\inst{1,2} \and Zhenbo Yu\inst{3} \and Xingyu Li\inst{1} \and \\  Bingbing Ni\inst{2,3} \and Minglun Gong\inst{4} \and Li Cheng\inst{1} }
\authorrunning{J. Yang et al.}
%
\institute{University of Alberta \and Huawei Hisilicon  \and Shanghai Jiao Tong University \and University of Guelph\\
\email{\{jyang7,xzuo,sen9,xingyu,lcheng5\}@ualberta.ca}}
\maketitle

\begin{abstract}
Given a single image of a general object such as a chair, could we also restore its articulated 3D shape similar to human modeling, so as to animate its plausible articulations and diverse motions?
This is an interesting new question that may have numerous downstream augmented reality and virtual reality applications. 
Comparing with previous efforts on object manipulation, our work goes beyond 2D manipulation and rigid deformation, and involves articulated manipulation.
To achieve this goal, we propose an automated approach to build such 3D generic objects from single images and embed articulated skeletons in them. Specifically, our framework starts by reconstructing the 3D object from an input image. Afterwards, to extract skeletons for generic 3D objects, we develop a novel skeleton prediction method with a multi-head structure for skeleton probability field estimation by utilizing the deep implicit functions. A dataset of generic 3D objects with ground-truth annotated skeletons is collected.
Empirically our approach is demonstrated with satisfactory performance on public datasets as well as our in-house dataset; 
our results surpass those of the state-of-the-arts by a noticeable margin on both 3D reconstruction and skeleton prediction. 
\keywords{Object Reconstruction, Object Rigging}
\let\thefootnote\relax\footnote{* equal contribution}
\let\thefootnote\relax\footnote{Project webpage: \url{https://kulbear.github.io/object-wakeup/}}

\end{abstract}

\section{Introduction}
\begin{figure}
    \centering
    \includegraphics[width=0.75\columnwidth]{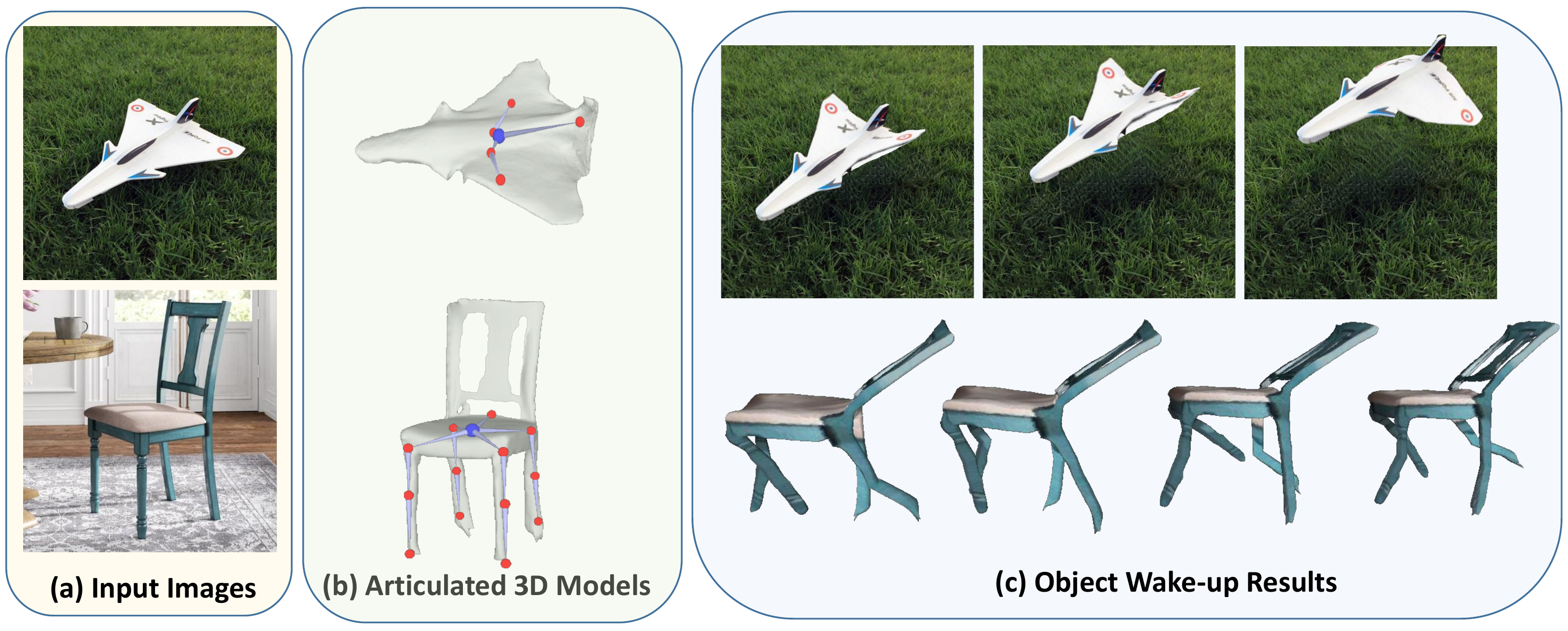}
    \caption{Two exemplar results. Given an input image of airplane or chair, our approach is capable of reconstructing its 3D shape with embedded skeleton and finally animating plausible articulated motions. 
    }
    \vspace{-5mm}
    \label{fig:titleImage}
\end{figure}

Presented with a single image of a generic object, say an airplane or a chair, our goal is to restore its 3D shape with the embedded skeleton. With the rigged 3D models, we can manipulate the object and generate its plausible articulations and possibly fun motions, such as an airplane flapping its wings or a chair walking as a quadruped, as illustrated in Fig.~\ref{fig:titleImage}. This new question considered in this paper essentially entails the extraction and manipulation of objects from images, which could have many downstream applications in virtual reality or augmented reality scenarios. 
It is worth noting that there has been research efforts~\cite{kholgade20143d} performing 3D manipulations from a single input image, where the main focus is on rigid transformations. To create non-rigid deformations, professional software has been relied on with intensive user interactions. Instead, we aim to automate the entire pipeline of object reconstruction, rigging, and animation.
The objects, as we considered here, are articulated -- objects that are capable of being controlled by a set of joints. 
In a sense, our problem could be considered as a generalization of image-based 3D human shape and pose reconstruction to generic objects encountered in our daily life, as long as they could be endowed with a skeleton.
\vspace{-10pt}

Compared with the more established topic of human shape and pose estimation~\cite{weng2019photo}, there are nevertheless new challenges to tackle with. To name one, there is no pre-existing parametric shape model for general objects. Besides, the human template naturally comes with its skeletal configuration for 3D motion control, and the precise skinning weights designed by professionals. However, such skeletal joints are yet to be specified not to mention the skinning weights in the case of generic objects, which usually have complex and diverse structures. 

To address those problems and restore 3D rigged models from an input image, building upon the achievements on 3D object modeling, we propose a stage-wise framework consisting of two major steps.
Step one involves 3D shape reconstruction from a single image.
We improve the baseline method by incorporating a transformer-based~\cite{pmlr-v139-touvron21a} encoder as the feature extractor, as well as an auxiliary voxel prediction module with improved loss function~\cite{mescheder2019occupancy}. With the reconstructed 3D model, in this paper, we focus more on the second step of predicting both the skeletal joints and bones from those 3D models. We propose a novel skeleton prediction method and reformulate it as estimating a multi-head probability field, inspired by the deep implicit functions of~\cite{mescheder2019occupancy}. Specifically, compared with previous skeleton prediction methods with voxel-based~\cite{xu2019predicting} or mesh-based representations~\cite{xu2020rignet}, we are able to predict the existence probability of joints and bones in a continuous 3D space.
To further improve the performance, a joint-aware instance segmentation is proposed and incorporated as an auxiliary task that considers regional features of neighboring points.

Our major contributions are listed as follows: 1) a new object wake-up problem is considered. For which an automated pipeline is proposed to restore 3D objects with embedded skeletons from single images. To our knowledge, it is the first attempt to deform and articulate generic objects from images; 
2) A novel and effective skeleton prediction approach with a multi-head structure is developed by utilizing the deep implicit functions.  
3) Moreover, in-house datasets (SSkel \& ShapeRR) of general 3D objects are constructed, containing annotated 3D skeletal joints and photo-realistic re-rendered images, respectively.
Empirically our entire pipeline is shown to achieve satisfactory results. 
Further evaluations on the public benchmarks and on our in-house datasets demonstrate the superior performance of our approach on the related tasks of image-based shape reconstruction and skeleton prediction. 


\begin{figure*}[h]
    \vspace{-4mm}
    \centering
    \includegraphics[width=1\textwidth]{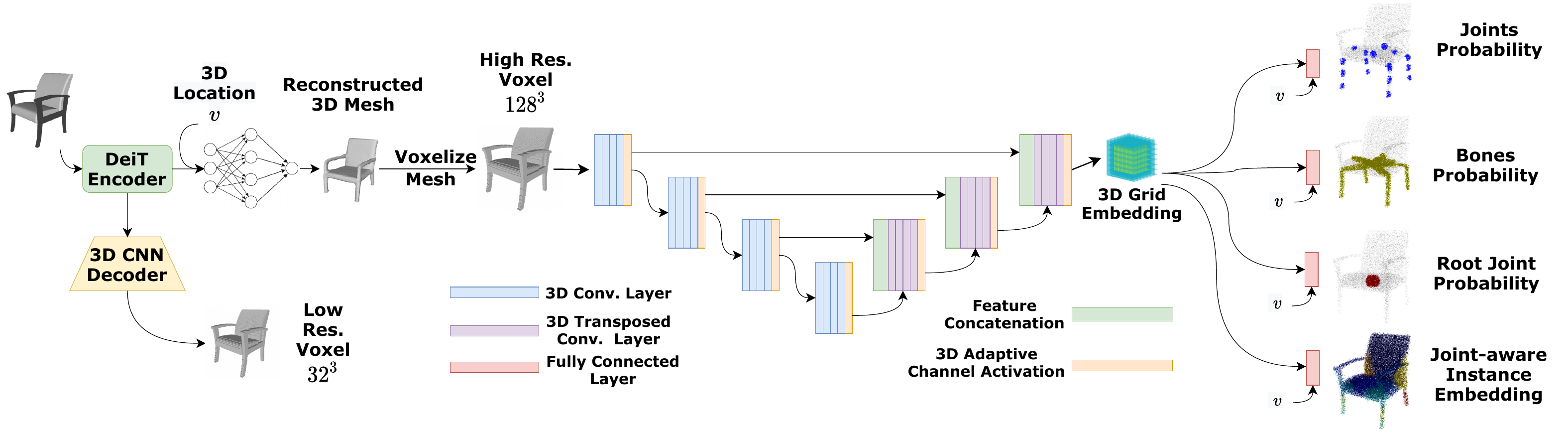}
    \vspace{-5mm}
    \caption{Our overall pipeline. It starts with the proposed Transformer-based model for 3D reconstruction consisting of a DeiT image encoder, an auxiliary 3D CNN voxel prediction branch and the occupancy decoder. The proposed SkelNet accepts the high resolution voxelized input from the reconstructed 3D mesh, and predicts articulated skeleton with a multi-head architecture.}
    \label{fig:overall}
    \vspace{-5mm}
\end{figure*}

\vspace{-3mm}
\section{Related Work}

\textbf{Image-based Object Reconstruction}. 
There exist numerous studies on image-based 3D object reconstruction with various 3D shape representations, including voxel, octree~\cite{riegler2017octnet,tatarchenko2017octree,wang2020deepoct}, deep implicit function, mesh and point cloud~\cite{fan2017point,lin2018learningpoint,qi2017pointnetplus,mi2020ssrnet}. Methods based on different representations have their own benefits and shortcomings. For example, as a natural extension of 2D pixels, voxel representation~\cite{gkioxari2019meshrcnn,tulsiani2018multi} has been widely used in early efforts due to its simplicity of implementation and compatibility with the convolutional neural network. However, these approaches often yield relatively coarse results, at the price of significant memory demand and high computational cost.
Mesh-based representations~\cite{kato2018neural,liu2019soft,wang2018pixel2mesh,kulon2020weaklymesh}, on the other hand, become more desirable in real applications, as they are able to model fine shape details, and are compatible with various geometry regularizers. It is however still challenging to work with topology changes~\cite{wang2018pixel2mesh,pan2019TMnet}.
Deep implicit 3D representations~\cite{park2019deepsdf,chen2019implicitfields,lin2020sdfsrn,tretschk2020patchnets} have recently attracted wide attention as a powerful technique in modeling complex shape topologies at arbitrary resolutions.

\textbf{Skeleton Prediction and Rigging}. 
The task of skeleton prediction has been investigated in various fields and utilized in a variety of applications for shape modeling and analysis. 
The best-known example is the medial axis~\cite{amenta1999surface,attali1997computing}, which is an effective means for shape abstraction and manipulation.
Curve skeleton or meso-skeleton~\cite{huang2013l1,yin2018p2p} have been popular in computer graphics, mostly due to their compactness and ease of manipulation.
It is worth noting the related research around detecting 3D keypoints from input point clouds, such as skeleton merger~\cite{ShiEtAl:CVPR21}.

Pinocchio~\cite{BarPop:siggraph07} is perhaps the earliest work on automatic rigging, which fits a pre-defined skeletal template to a 3D shape, with skinning obtained through heat diffusion. These fittings, unfortunately, tend to fail as the input shapes become less compatible with the skeletal template. On the other hand, hand-crafting templates for every possible structural variation of an input character is cumbersome. More recently, Xu et al.\cite{xu2019predicting} propose to learn a volumetric network for inferring skeletons from input 3D characters, which however often suffers from the limited voxel resolution. Exploiting the mesh representation, RigNet~\cite{xu2020rignet} utilizes a graph neural network to produce the displacement map for joint estimation, which is followed by the additional graph neural networks to predict joint connectivity and skinning weights. Its drawback is they assume strong requirements for the input mesh such as a watertight surface with evenly distributed vertices can be satisfied. Besides, they predict the joints and kinematic chains successively causing error propagation from stages.. 
In contrast, a deep implicit function representation~\cite{mescheder2019occupancy} which is capable of predicting the joints and bones over a continuous 3D space is considered in this paper for inferring skeleton.

\textbf{Image based Object Animation}. 
An established related topic is photo editing, which has already been popular with professional tools such as PhotoShop. Existing tools are however often confined to 2D object manipulations in performing basic functions such as cut-and-paste and hole-filling.
A least-square method is considered in~\cite{schaefer2006image} to affine transform objects in 2D.
The work of~\cite{hornung2007character} goes beyond linear transformation, by presenting an as-rigid-as-possible 2D animation of a human character from an image, it is however manual intensive.
In~\cite{xu2008animating}, 2D instances of the same visual objects are ordered and grouped to form an instance-based animation of non-rigid motions.
Relatively few research activities concern 3D animations, where the focus is mostly on animals, humans, and human-like objects. For example, photo wake-up~\cite{weng2019photo} considers reconstruction, rig, and animate 3D human-like shapes from input images. This line of research benefits significantly from the prior work establishing the pre-defined skeletal templates and parametric 3D shape models for humans and animals. 
On the other hand, few efforts including~\cite{kholgade20143d,chen20133} consider 3D manipulations of generic objects from images, meanwhile, they mainly focus on rigid transformations.
Our work could be regarded as an extension of automated image-based human shape reconstruction \& animation to reconstruct \& articulate generic lifeless objects from single images.


\vspace{-4mm}
\section{Our Approach}
\vspace{-1mm}
Given an input image, usually in the form of a segmented object, first the 3D object shape is to be reconstructed; its skeletons are then extracted to form a rigged model. In this section, we will present the stage-wise framework in detail.


\subsection{Image-based 3D Shape Reconstruction}
A Transformer-based occupancy prediction network is developed here, which performs particularly well on real images when compared with existing methods~\cite{mescheder2019occupancy,xu2019disn,li2021d2im}.
As illustrated in Fig.~\ref{fig:overall}, it consists of a 2D transformer encoder, an auxiliary 3D CNN decoder, and an occupancy decoder. 
The DeiT-Tiny~\cite{pmlr-v139-touvron21a} is used as our transformer encoder network. Similar to the Vision Transformer~\cite{dosovitskiy2020image}, the encoder first encodes fixed-size patches splitted from the original image and processes extract localized information from each of the patches, then outputs a universal latent representation for the entire image by jointly learning the patch representation with multi-head attention. An auxiliary 3D CNN decoder is used for reconstructing a low-resolution voxel-based 3D model as well as helping to encode 3D information for the latent representation extracted from the Transformer encoder. The occupancy decoder then uses the latent representation as the conditional prior to predict the occupancy probability for each point by introducing fully connected residual blocks and conditional batch normalization~\cite{peng2020convolutional,niemeyer2020differentiable}.


It is worth noting that although the voxel prediction branch is only used for auxiliary training, the highly unbalanced labels where most of the voxel occupancy are zeros will always make the training more difficult. To this end, while most of the methods for voxel-based 3D reconstruction simply use the (binary) cross-entropy loss which is directly related to IoU metric~\cite{shi20213d}, in this work, the Dice loss is extended to gauge on both the 3D voxel prediction and the point-based occupancy prediction,
\begin{equation}
\label{eq:jointsL}
\mathcal{L}_{dice} = 1 - \frac{\sum^{N^3}_{n=1}\hat{y}_{n}y_{n}}{\sum^{N^3}_{n=1}\hat{y}_{n} + y_{n}} - \frac{\sum^{N^3}_{n=1}(1-\hat{y}_{n})(1-y_{n})}{\sum^{N^3}_{n=1}2 - \hat{y}_{n} - y_{n}},
\end{equation}
where $y_{n}$ is the ground-truth occupancy score, $\hat{y}_{n}$ is the predicted occupancy score of the $n$-th element.

\subsection{Skeleton Prediction and Automatic Rigging}

Our key insight here is instead of predicting the joints inside fixed voxel locations~\cite{xu2019predicting} or indirectly regressing the joints location by estimating the displacement from the mesh~\cite{xu2020rignet}, we train a neural network utilizing the deep implicit function to assign every location with a probability score in $[0, 1]$, indicating the existence of a skeletal joint and bone. Taking the 3D model and any sampled 3D point location as input, the network produces the joint and bone existence probabilities. 
In addition, we incorporate joint-aware instance segmentation as an auxiliary task considering the regional features over neighboring points. In inference, the feature embedding output from the instance segmentation branch is further used in the subsequent step to infer joint locations from the incurred joints' probability maps. 


As in Fig.~\ref{fig:overall}, four output heads are utilized, which are for predicting the probability of skeletal joints, the root joint, the bones, and the joint-aware instance segmentation, respectively. The output from the instance segmentation is a feature embedding.


\begin{wrapfigure}{r}{0.5\textwidth}
\begin{center}
    \includegraphics[width=0.48\textwidth]{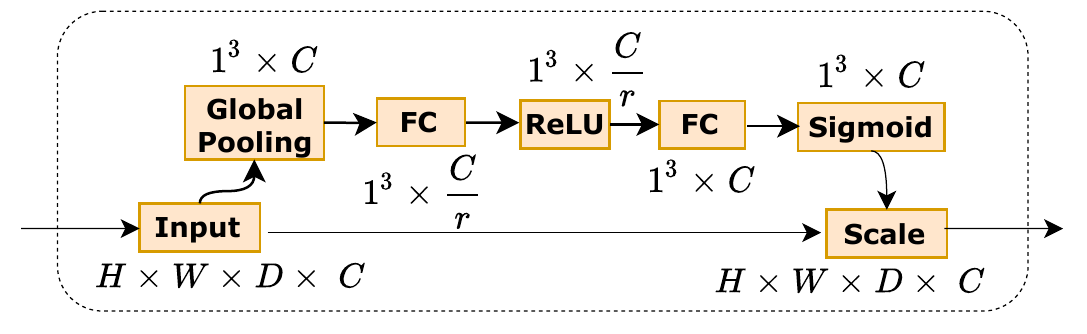}
    \end{center}
    \vspace{-5mm}
    \caption{The 3D adaptive channel activation module.}
    \vspace{-5mm}
    \label{fig:3dadp}
\end{wrapfigure}
\textbf{Feature Extraction}. 
The predicted 3D shape, represented as an occupancy grid with the dimension of 128$^{3}$, is converted to a 3D feature embedding grid by a 3D UNet structure. Inspired by the design of Squeeze and Excitation (SE) block in 2D image classification, a 3D adaptive channel activation module is developed as a plug-in module, to be attached after each of the encoder and decoder blocks of the 3D UNet, as shown in Fig.~\ref{fig:3dadp}.
The empirical ablative study demonstrated the usefulness of this 3D adaptive channel activation module.


\textbf{Multi-head Implicit Functions}. 
Given aggregated features from the feature extraction, we acquire the feature vector for any 3D point $v$ via the trilinear interpolation from 3D feature embedding. 
For each of the output heads, a fully-connected network (empirically it is implemented as 5 fully-connected ResNet blocks and ReLU activation~\cite{peng2020convolutional,niemeyer2020differentiable}) is engaged to take as input the point $v$ and its feature vector. 
The concurrent multi-head strategy eliminates the possible issue with error propagation of successive prediction~\cite{xu2020rignet}. 

\textbf{Sampling}. In general, the animation joints and bones should lie inside the convex hull of the object. Therefore, different from previous efforts that uniformly sample points in a 3D volume~\cite{mescheder2019occupancy,peng2020convolutional}, points in our 3D space are adaptively sampled. Specifically, for each sample in the training batch, we sampled $K$ points with 10$\%$ of the points lying outside but near the surface, and the rest 90$\%$ points entirely inside the object. 

\textbf{Joints and Bones Loss}.
First, for every query point, its joint probability is computed under a 3D Gaussian distribution measured by its distance to nearest annotated joint locations. 
To generate the bone probability field, for every query point we compute a point-to-line distance to its nearest line segment of the bones, and the bone probability is computed under the Gaussian distribution of the measured distance. 
In training, with the query points $v\in \mathbb{R}^{3}$ acquired through sampling, the network predicts their probabilities of being a joint or lying on bones. 
Different from the occupancy prediction~\cite{mescheder2019occupancy} task where the binary cross-entropy loss is used, we use the L1 loss to measure the difference of the predicted joint probability and their ground-truth values as we are dealing with the continuous probability prediction:
for the $i$-th sample in training, the loss function is defined as,
\vspace{-2mm}
\begin{equation}
\label{eq:jointsL}
\begin{aligned}
\mathcal{L}_{joint}^{i}(\hat{P_J}, P_J) & = \sum_{v \in \mathcal{V}^i} |\hat{P_J}(v) - P_J(v)| \\
\mathcal{L}_{jointR}^{i}(\hat{P_{JR}}, P_{JR}) & = \sum_{v \in \mathcal{V}^i} |\hat{P_{JR}}(v) - P_{JR}(v)|
\end{aligned}
\end{equation}
\vspace{-2mm}

In the above equation, $\hat{P_J}$ is the predicted joints probability field, and $P_J$ is the ground-truth probability field. $\hat{P_{JR}}$ and $P_{JR}$ denote for the probability field of the root joint. $\mathcal{V}^i$ denotes the sampled points for the $i$-th model.

Similarly, for the sampled points, L1 loss is also applied between predicted bones probability $\hat{P_B}$ and the ground-truth $P_B$. The loss function of the bones is denoted as $\mathcal{L}_{bone}^{i}(\hat{P_B}, P_B)$.


\textbf{Symmetry Loss}.
Since the objects of interest often possess symmetric 3D shapes, a symmetry loss is used here to regularize the solution space, as follows, 
\vspace{-2mm}
\begin{equation}
\label{eq:sym}
\begin{aligned}
\mathcal{L}_{symJ}^{i}(\hat{P_J}) & = \textbf{1}_{\Omega'}(i) \sum_{v \in \mathcal{V}^i}|\hat{P_J}(v) - \hat{P_J}(\phi(v))|, \\
\mathcal{L}_{symB}^{i}(\hat{P_B}) & = \textbf{1}_{\Omega'}(i) \sum_{v \in \mathcal{V}^i}|\hat{P_B}(v) - \hat{P_B}(\phi(v))|,
\end{aligned}
\end{equation}
\vspace{-2mm}

Here $\phi(v)$ denotes the mapping from point $v$ to its symmetric point. To detect the symmetry planes, as the input 3D mesh models are in the canonical coordinates, we flip the mesh model according to the xy-, xz- and yz-planes. The symmetry plane is set as the one with the smallest Chamfer distance computed between the flipped model and the original model. $\textbf{1}_{\Omega'}$ is an indicator function where $\Omega'$ is the subset of training models with symmetry planes detected.

\textbf{Joint-aware Instance Segmentation Loss}. 
The joint-aware instance segmentation maps the sampled point from Euclidean space to a feature space, where 3D points of the same instance are closer to each other than those belonging to different instances. To maintain consistency between the clustered feature space and the joints probability maps, the part instance is segmented according to the annotated ground-truth joints. Basically, for each sampled point we assign an instance label as the label or index of its closest joint. 
Following the instance segmentation method of~\cite{wang2018sgpn}, our joint-aware instance segmentation loss is defined as a weighted sum of three terms: 
(1) $\mathcal{L}_{\textit{var}}$ is an intra-cluster variance term that pulls features belonging to the same instance towards the mean feature; (2) $\mathcal{L}_{\textit{dist}}$ is an inter-cluster distance term that pushes apart instances with different part labels; and (3) $\mathcal{L}_{\textit{reg}}$ is a regularization term that pulls all features towards the origin in order to bound the activation.
\vspace{-2mm}
{
\begin{align}
\label{eq:inst-var}
\nonumber
\mathcal{L}_{\textit{var}}^{i}(\mu,x) & = \frac{1}{|J^i|} \sum_{c=1}^{|J^i|}\frac{1}{N_c}\sum_{j=1}^{N_c}[\ \| \mu_c^i - x_j^i \| - \delta_{\textit{var}} ]\ _{+}^{2}, 
\\ 
\nonumber
\mathcal{L}_{\textit{dist}}^{i}(\mu) & = \frac{1}{|J^i|(|J^i|-1)} \sum_{c_{a}=1}^{|J^i|}\sum_{\substack{c_{b}=1 \\ c_{b}\neq c_{a}}}^{|J^i|}[\ 2\delta_{\textit{dist}} - \| \mu_{c_a}^i - \mu_{c_b}^i \| ]\ _{+}^{2}, 
\\ 
\mathcal{L}_{\textit{reg}}^{i}(\mu) & = \frac{1}{|J^i|} \sum_{c=1}^{|J^i|} \| \mu_{c}^i \|.
\end{align}
}

Here $|J^i|$ denotes the number of joints or clusters for the $i$-th sample model. $N_c$ is the number of elements in cluster $c$. $x_j^i$ is the output feature vector for the query point. $[\ x]\ _{+}$ is the hinge function. The parameter $\delta_{\textit{var}}$ describes the maximum allowed distance between a feature vector and the cluster center. Likewise, $2\delta_{\textit{dist}}$ is the minimum distance between different cluster centers to avoid overlap.

\textbf{Joints and Kinematic Tree Construction}. 
In inference, the joints and bones are obtained from the corresponding probability maps by mean-shift clustering. 
Instead of clustering over the euclidean space as in classical mean-shift clustering, we implement the clustering on the feature space with the kernel defined over the feature embedding output from the joint-aware instance segmentation. In this way, the points belonging to the same joint-aware instance will all shift towards the corresponding joints. The kernel is also modulated by the predicted joint probability to better localize the joint location. Mathematically, at each mean-shift iteration, for any point $v$ it is displaced according to the following vector:
\vspace{-2mm}
\begin{equation}
\label{eq:ms}
m(v) = \frac{\sum_{u \in \mathcal{N}(v)} P_J(v)\kappa(\| x(u)-x(v)\|)u}{\sum_{u \in N(v)} P_J(v)\kappa(\| x(u)-x(v)\|)} - v
\end{equation}
where $\mathcal{N}(v)$ denotes the neighboring points of $v$, $x(v)$ is the feature embedding output from our joint-aware instance segmentation. Besides, $\kappa()$ is a kernel function and in our case we choose to use the RBF kernel.
%
Following~\cite{xu2019predicting}, the object kinematic tree (or chains) are constructed using a minimum spanning tree by minimizing a cost function defined over the edges connecting the joints pair-wisely. 
It is realized as a graph structure, with the detected joints as the graph nodes, and the edges connecting the pairwise joints computed from the probability maps.
Specifically, for every edge, its weight is set by the negative-log function of the integral of the bones probability for the sampled points over the edge. The MST problem is solved using Prim's algorithm~\cite{cheriton1976finding}.

\textbf{Skinning Weight Computation.}
For automatic rigging of the reconstructed 3D model, the last issue is to compute the skinning weights that bind each vertex to the skeletal joints. 
To get meaningful animation, instead of computing the skinning weights according to the Euclidean distance~\cite{BarPop:siggraph07}, we choose to assign the skinning weights by utilizing the semantic part segmentation~\cite{wang2018sgpn}. Specifically, for every segmented part, we assign its dominant control joint to the one closest to the center of the part. In some cases where the center of the part could have about the same distance to more than one skeletal joint, we choose the parent joint as the control joint. The skinning weights around the segmentation boundaries are smoothed out afterwards. It is worth noting that some semantic parts are further segmented if skeleton joints are detected inside the part.

\subsection{Our In-house Datasets}
As there is no existing dataset of general 3D objects with ground-truth skeletons, we collect such a dataset (named SSkel for \emph{ShapeNet skeleton}) by designing an annotation tool to place joints and build kinematic trees for the 3D shapes. To ensure consistency, a predefined protocol is used for all object categories. For example, for chairs, we follow the part segmentation in PartNet dataset~\cite{mo2019partnet} to segment a chair into the chair seat, back, and legs. The root joint is annotated at the center of the chair seat, followed by child joints which are the intersection between chair seat and back, chair seat and legs. More details about the annotation tool and some sampled annotations are presented in the supplementary. Without loss of generality, we only consider four categories of objects from ShapeNet~\cite{chang2015shapenet}, namely \emph{chair}, \emph{table}, \emph{lamp} and \emph{airplane}. Our SSkel dataset contains a total of 2,150 rigged 3D shapes including 700 for chair, 700 for table, 400 for lamp and 350 for airplane.

Moreover, in improving the input image resolution and quality of the original ShapeNet, we use the UNREAL 4 Engine to re-render photo-realistic images of the 3D ShapeNet models with diverse camera configuration, lighting conditions, object materials, and scenes, named ShapeRR dataset for \emph{ShapeNet of realistic rendering}. More details are relegated to the supplementary file.

\section{Experiments}

 

\textbf{Datasets}. 
A number of datasets are considered in our paper. In terms of image-based reconstruction, it contains our ShapeRR dataset for synthetic images and the Pix3D dataset of real images. In terms of rigging performance, we use the RigNetv1 dataset for 3D shape-based rigging, and our SSkel dataset for image-based rigging.
The Pix3D dataset contains 3D object shapes aligned with their real-world 2D images. 
Similar to ShapeRR, 
we focus on a subset of 4 categories in the dataset, i.e. chair, sofa, desk, and table. 
The RigNetv1 dataset (i.e. ModelsResource-RigNetv1~\cite{xu2019predicting}), on the other hand, contains 2,703 rigged 3D characters of humanoids, quadrupeds, birds, fish, robots, and other fictional characters.





\subsection{Evaluation on Image-based Reconstruction}
For evaluation metrics, we follow the previous works~\cite{mescheder2019occupancy} and use volumetric IoU and Chamfer-L1 distance. 
We first compare with several state-of-the-art methods with released source code on single image object reconstruction where each of the methods is trained and tested, namely OccNet~\cite{mescheder2019occupancy}, DVR~\cite{niemeyer2020differentiable} and D$^{2}$IM-Net~\cite{li2021d2im}. We follow the common test protocol on ShapeNet as it has been a standard benchmark in the literature. All methods are re-implemented (when the code is not available) and re-trained then evaluated directly on the test split. We can observe that our method performs reasonably well compared with other recent methods, and outperforms existing methods in 3 of the 4 categories. And we are able to achieve a significant advantage over other methods in terms of the average performance across all 4 categories of our interests.

\begin{figure}[h]
    \centering
    \includegraphics[width=0.95\columnwidth]{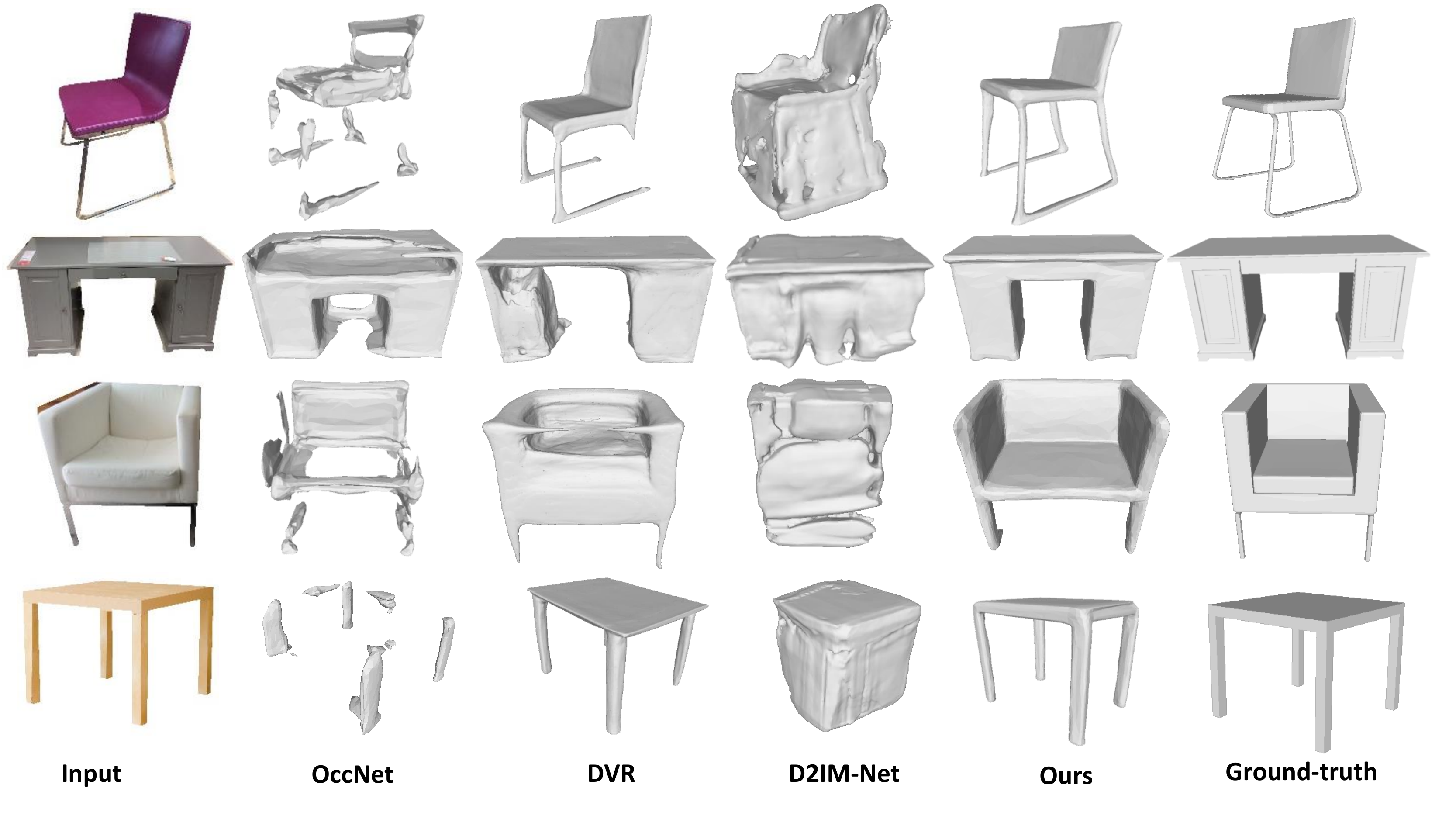}
    \caption{Visualization of image-based 3D reconstruction on the Pix3D dataset. Our method shows excellent generalization performance on the real images.}
    \label{fig:pix_cmp}
\end{figure}

Considering that our 3D reconstruction is primarily for supporting rigging and animation purposes on real images, to better compare the generalization ability with such a situation, we use the complete Pix3D dataset as the test set. 
We report both quantitative and visual comparison on Pix3D in Tab.~\ref{tab:quantitative} and in Fig.~\ref{fig:pix_cmp} respectively. As shown in Tab.~\ref{tab:quantitative}, our proposed method has outperformed all previous approaches on Pix3D with a large margin in terms of the two metrics. To validate the effectiveness of our feature encoder and the incorporated auxiliary voxel prediction task, we also conduct a group of ablative studies, and the experiment results are included in the supplementary material. 
\vspace{-2mm}
\begin{table*}[!]
	\centering
    \resizebox{\textwidth}{!}{
    \begin{tabular}{lcccccccccc}
    \hline
    & \multicolumn{5}{c}{---------------~Chamfer Distance~($\downarrow$)~---------------} & \multicolumn{5}{c}{--------------~Volumetric IoU~($\uparrow$)~--------------} \\
    ShapeNet & Chair  & Table  & Lamp  & Airplane & Avg.  & Chair  & Table  & Lamp  & Airplane & Avg. \\ \hline
    OccNet~\cite{mescheder2019occupancy} & 1.9347    &  1.9903   &  4.5224   &  1.3922   & 2.3498   & 0.5067   & 0.4909  & 0.3261 & 0.5900 & 0.4918   \\
    DVR~\cite{niemeyer2020differentiable}  & 1.9188    &  2.0351   &  4.7426   &  1.3814   & 2.5312   &   0.4794  &  0.5439   & 0.3504 &  0.5741  & 0.5029  \\
    D$^{2}$IM-Net~\cite{li2021d2im} &  \textbf{1.8847}   &    1.9491 &   4.1492 & 1.4457     & 2.0346   & \textbf{0.5487}   & 0.5332   &  0.3755 &  0.6123 &  0.5231   \\
    Ours     & 1.8904  & \textbf{1.7392} & \textbf{3.9712}    &  \textbf{1.2309}    &  \textbf{1.9301}   &  0.5436   & \textbf{0.5541
} & \textbf{0.3864}  & \textbf{0.6320}  &  \textbf{0.5339}  \\ \hline
    
    Pix3D & Table  & Chair  & Desk  & Sofa & Avg.  & Table  & Chair  & Desk  & Sofa & Avg. \\ \hline
    
    OccNet~\cite{mescheder2019occupancy} & 7.425    &  9.399   &  15.726   &  14.126   & 11.625   & 0.215   & 0.201  & 0.143 & 0.152 & 0.190   \\
    DVR~\cite{niemeyer2020differentiable}  & 8.782    &  6.452   &  12.826   &  11.543   & 9.901   &   0.187  &  0.237   & 0.165 &  0.187  & 0.185  \\
    D$^{2}$IM-Net~\cite{li2021d2im} &  8.038   &   7.592  &  11.310 & 9.291    &  9.057  & 0.205   & 0.244   &  0.183 &  0.207 &  0.215   \\
    Ours     & \textbf{6.449}  & \textbf{6.028} & \textbf{8.452}    &  \textbf{8.201}    &  \textbf{7.282}   &  \textbf{0.239}   & \textbf{0.277
} & \textbf{0.219}  & \textbf{0.241}  &  \textbf{0.242}  \\  \hline
    \end{tabular}}
\caption{Image-based 3D mesh reconstruction on ShapeRR (i.e. re-rendered ShapeNet dataset) and Pix3D dataset. Metrics are Chamfer Distance (×0.001, the smaller the better) and Volumetric IoU (the larger the better). Best results are in \textbf{bold face}.}
\label{tab:quantitative}
\vspace{-8mm}
\end{table*}

\vspace{-3mm}
\vspace{-5mm}


\subsection{Evaluation on Skeleton Prediction}
The evaluation is conducted on both the RigNetv1 dataset and our SSkel dataset, where our approach is compared with two state-of-the-art methods, RigNet~\cite{xu2020rignet} and VolumetricNets~\cite{xu2019predicting}. 

\textbf{Metrics.} 
First, we measure the accuracy of the predicted joints by computing the Chamfer distance between the predicted joints and the ground-truth which is denoted as CD-J2J. Similarly, the predicted bones are evaluated by computing the Chamfer distance between the densely sampled points over the estimated bones and the ground-truth, which is denoted as CD-B2B. CD-J2B is also considered here by computing the Chamfer distance between predicted joints to bones. For all metrics, the lower the better.


\textbf{Quantitative evaluation.}
In Tab.~\ref{tab:comparerignet} we show the comparison results of the predicted skeleton on the RigNetv1 dataset~\cite{xu2019predicting}. For the RigNetv1 dataset, we follow the same train and test split as previous works~\cite{xu2019predicting,xu2020rignet}. 
In Tab.~\ref{tab:comparerignet} we show the quantitative evaluation and comparison results of the predicted skeleton on our SSkel dataset. We have re-trained the RigNet~\cite{xu2020rignet}, which is the most current work on auto-rigging, on our SSkel dataset. As shown in the tables, our proposed skeleton prediction method has outperformed the current state-of-the-art approaches with the smallest error on all reported metrics on both the RigNetv1 dataset and our SSkel dataset.

It is worth noting that the evaluation on the SSkel dataset is conducted with two different inputs. First, we report the skeleton error(RigNet-GT, Ours-GT) when taking the ground-truth 3D models as input. To evaluate the performance of the overall pipeline, we also calculate the skeleton error(RigNet-rec, Ours-rec) when 3D models reconstructed from the color images are taken as input. Our skeleton prediction performance on the reconstructed 3D models degraded slightly due to imperfect reconstruction.

\vspace{-3mm}
\begin{table}[h]
\centering
\begin{tabular}{@{}cccc@{}}
\toprule
& CD-J2J ~($\downarrow$)~  & CD-J2B ~($\downarrow$)~              & CD-B2B ~($\downarrow$)~             \\ \midrule
Pinocchino~\cite{BarPop:siggraph07} & 0.072 & 0.055 & 0.047   \\
Volumetric~\cite{xu2019predicting} & 0.045 & 0.029 & 0.026  \\
RigNet~\cite{xu2020rignet}   & 0.039 & 0.024 & 0.022  \\
\textbf{Ours} & \textbf{0.029} & \textbf{0.019} & \textbf{0.017}  \\ \bottomrule
\end{tabular}
\caption{Comparison of skeleton prediction on the RigNetv1 dataset.}
\label{tab:comparerignet}
\vspace{-1mm}
\end{table}
\vspace{-4mm}

\vspace{-4mm}
\begin{table*}[!]
	\centering
    \centering
    \resizebox{\textwidth}{!}{
    \begin{tabular}{lccccccccccccccc}
    \hline
    & \multicolumn{3}{c}{--------Chair--------} & \multicolumn{3}{c}{-------Table-------}  & \multicolumn{3}{c}{--------Lamp--------} & \multicolumn{3}{c}{------Airplane------} & \multicolumn{3}{c}{------Average------}\\
    metrics & J2J & J2B & B2B  & J2J & J2B & B2B & J2J & J2B & B2B  & J2J & J2B & B2B  & J2J & J2B & B2B \\ \hline
    RigNet-GT & {0.052} & {0.042}  & {0.035} & {0.061}  & {0.049}  &  {0.040}    &  {0.132}  & {0.110}  &  {0.098}   & {0.096} & {0.081}  & {0.073}  & {0.061} & {0.046}  &  {0.041}     \\
    Ours-GT     & \textbf{0.030} & \textbf{0.023} & \textbf{0.021} & \textbf{0.044} & \textbf{0.032}   &  \textbf{0.028}    &  \textbf{0.097} &  \textbf{0.071}  &  \textbf{0.063}   & \textbf{0.075} & \textbf{0.062} & \textbf{0.056}  & \textbf{0.047} & \textbf{0.038} &  \textbf{0.033}\\
    RigNet-rec & {0.048} & {0.035}  & {0.033} & {0.060}  & {0.046}  &  {0.038}    &  {0.143}  & {0.116}  &  {0.102}   & {0.103} & {0.084}  & {0.076}  & {0.063} & {0.047} &  {0.042}     \\
    Ours-rec  & \textbf{0.036}  & \textbf{0.024} & \textbf{0.022}    &  \textbf{0.047}    &  \textbf{0.033}   &  \textbf{0.029}   & \textbf{0.101} & \textbf{0.073}  & \textbf{0.065}  &  \textbf{0.081} & \textbf{0.065} & \textbf{0.059} & \textbf{0.051} & \textbf{0.041} & \textbf{0.036}\\ \hline
    \end{tabular}}
\caption{Quantitative comparison of skeleton prediction on our SSkel dataset. The J2J, J2B, B2B are the abbreviation for CD-J2J, CD-J2B and CD-B2B respectively. For these values, the smaller the better. Best results are in \textbf{bold face}.}
\label{tab:quanShapeNet}
\vspace{-1mm}
\end{table*}
\vspace{-5mm}

\begin{figure}[!]
    \centering
    \includegraphics[width=0.85\columnwidth]{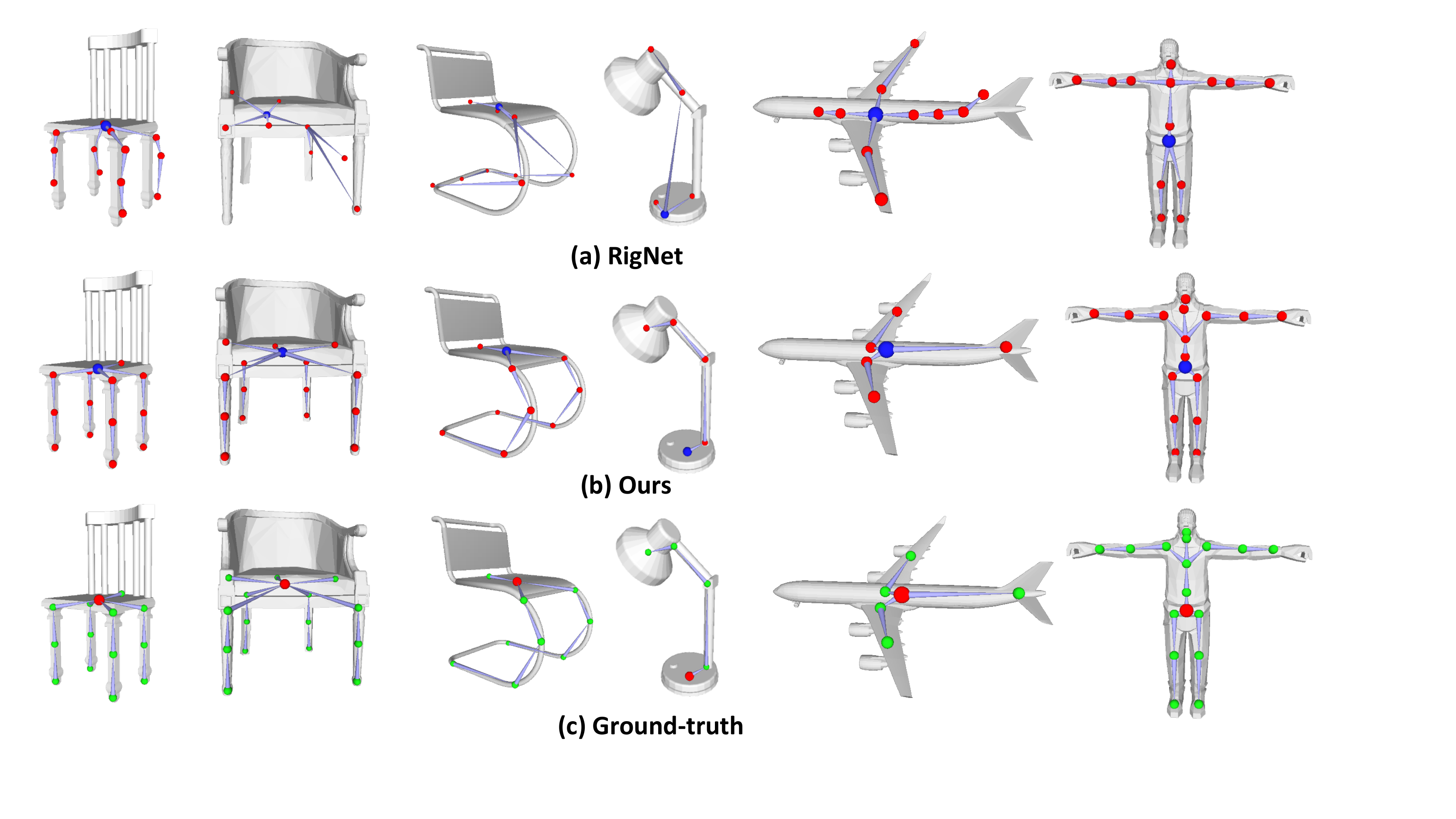}
    \caption{Visual comparison on skeleton prediction. The rightmost model comes from the RigNetv1 dataset and the others are from our SSkel dataset.}
    \label{fig:riggingCmp}
    \vspace{-1mm}
\end{figure}
\begin{figure}[!]
    \centering
    \includegraphics[width=0.80\columnwidth]{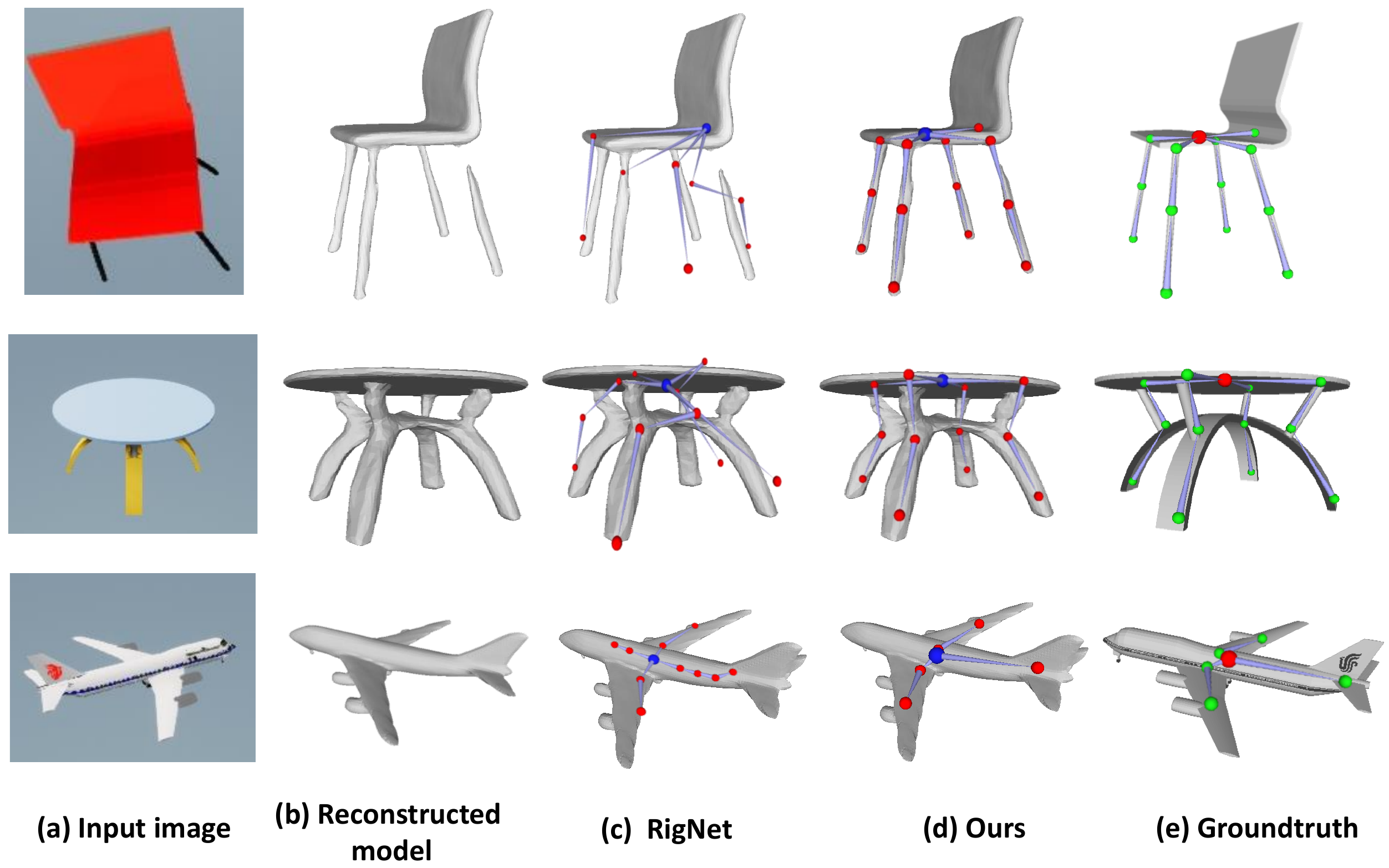}
    \caption{Visual results on articulated 3D models from input images. Taking the color image(a) as input, we reconstruct the 3D model(b) and predict its skeleton(d), and also compare with the RigNet~\cite{xu2020rignet} on skeleton prediction(c).}
    \label{fig:riggingCmp2}
    \vspace{-1mm}
\end{figure}

\textbf{Visual results on skeleton prediction.}
In Fig.~\ref{fig:riggingCmp} and Fig.~\ref{fig:riggingCmp2} we demonstrate the qualitative comparison of the predicted skeleton. First, in Fig.~\ref{fig:riggingCmp}, the skeletons are predicted with ground-truth 3D models as input. We also evaluated the overall pipeline when taking a single image as input, and the results are shown in Fig.~\ref{fig:riggingCmp2}.
As shown in the figures, compared with the most current work, our proposed approach can produce more reasonable results that correctly predicted the joints' location and constructed the kinematic chains. On the other hand, the RigNet method fails to localize the joints. The reason is that their mesh-based approach requires the vertices to be evenly distributed over the mesh and they rely on the mesh curvature to pre-train an attention model. But for the models from the SSkel dataset, there is no close connection between the mesh curvature and the joint locations. 
\vspace{-2mm}
\vspace{-2mm}
\vspace{-2mm}

\textbf{Ablation study.}
To validate the effectiveness of several key components of the proposed method, we conduct several ablation studies with the quantitative evaluation results shown in Table~\ref{tab:rigalbation}. We denote our method without the 3D channel adaptive activation, symmetry loss, and joint-aware instance segmentation as the Baseline method.
\vspace{-6mm}
\begin{table}[h]
\centering
\begin{tabular}{@{}ccc@{}}
\toprule
& RigNetv1  & SSkel 
\\ \midrule
Baseline                   & 0.037 & 0.065  \\
Baseline + joint-aware seg    & 0.033 & 0.055  \\
Baseline + symmetry loss   & 0.034 & 0.058  \\
Baseline + 3D adaptive activation        & 0.033 &  0.056  \\
Ours                       & \textbf{0.029} & \textbf{0.047}  \\ \bottomrule
\end{tabular}
\caption{Ablation study on joints prediction. CD-J2J metric is used.}
\vspace{-1mm}
\label{tab:rigalbation}
\end{table}
\vspace{-13mm}

\begin{figure*}[h]
    \centering
    \includegraphics[width=0.85\textwidth]{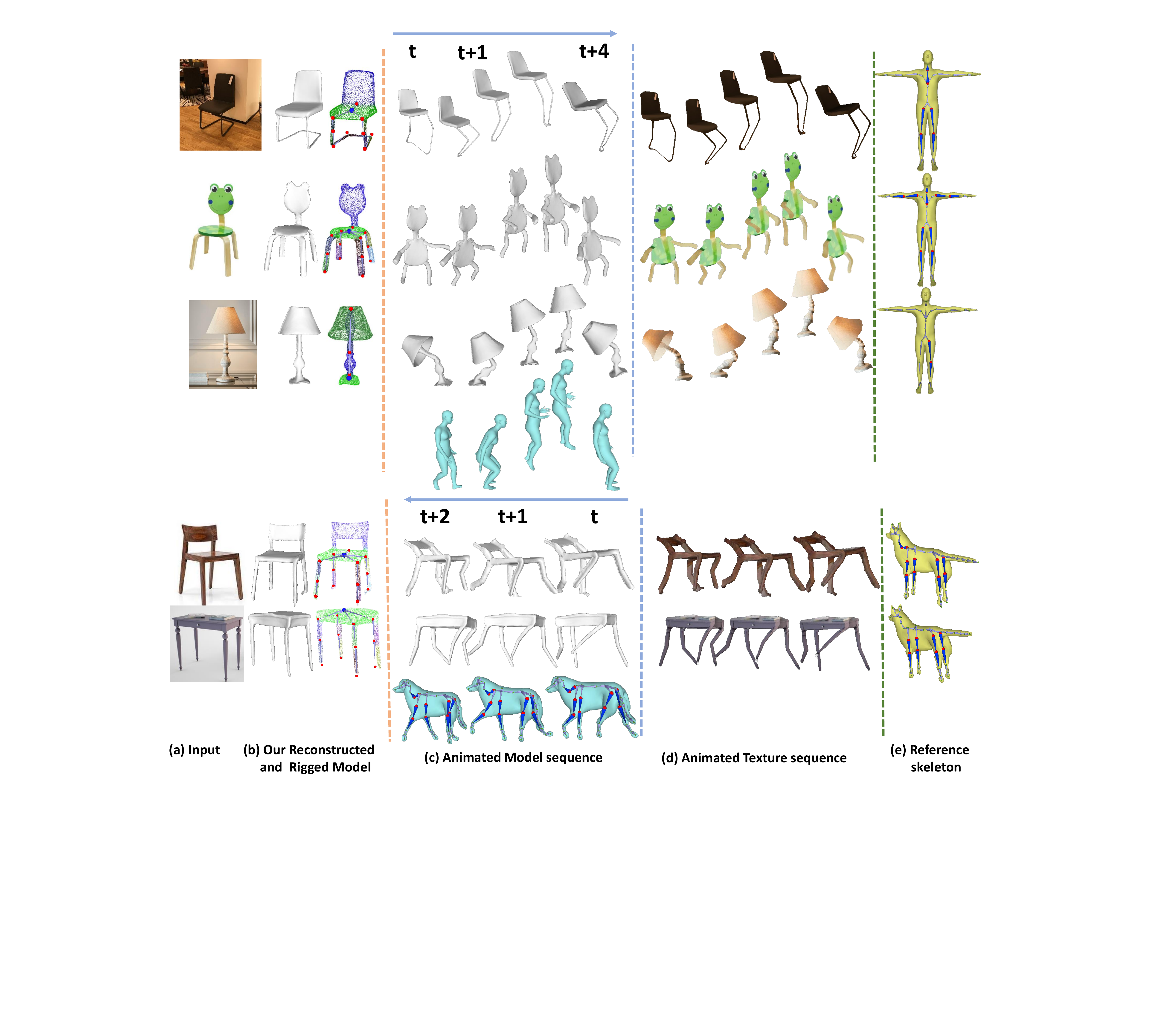}
    \caption{Object animation. Given an input image (i.e. the object segment), its 3D shape is reconstructed and rigged, followed by the animated sequence (re-targeted from human or quadruped motions, which is not the main focus of this work). We map the joints from the human or quadruped skeleton to the objects, and the mapped joints are marked in red (c). The source human/dog motion is shown in the bottom row.}
    \label{fig:animation_combine}
\vspace{-1mm}
\end{figure*}
\subsection{Applications on Animation}
After obtaining the rigged 3D models from the input images, in this section, we present interesting applications of animating the rigged 3D objects. To get the texture for the 3D models, similar to~\cite{xu2019disn} we have trained a deep neural network to predict the projection matrix represented as a 6D rotation vector aligning the 3D models from canonical space to image space. Our reconstructed 3D model is further refined and deformed according to the object silhouettes~\cite{weng2019photo}.
The mirrored texture is applied to the invisible part of the 3D model.

In Fig.~\ref{fig:animation_combine}, we demonstrate the animation of objects as driven by the source motion of reference articulated models. Specifically, in the upper rows of Fig.~\ref{fig:animation_combine} we map the motion of a Jumping human to two Chairs as well as a Lamp. The details of the skeleton mapping from the human template to the animated objects are shown in each corresponding row of Fig.~\ref{fig:animation_combine}(d). Likewise, in the lower part of Fig.~\ref{fig:animation_combine}, we demonstrate the manipulation of a Chair and Table driven by a quadruped. It is conducted by mapping the joints of four legs on the Dog skeleton to the legs of the chair and table. In addition, the joint of the neck is mapped to the joint on the chair back. The motion sequence of the dog is from RGBD-Dog dataset~\cite{kearney2020rgbd}. More results can be seen in the supplementary video.
\vspace{-2mm}
\begin{wrapfigure}{r}{0.475\textwidth}
\begin{center}
    \includegraphics[width=0.425\textwidth]{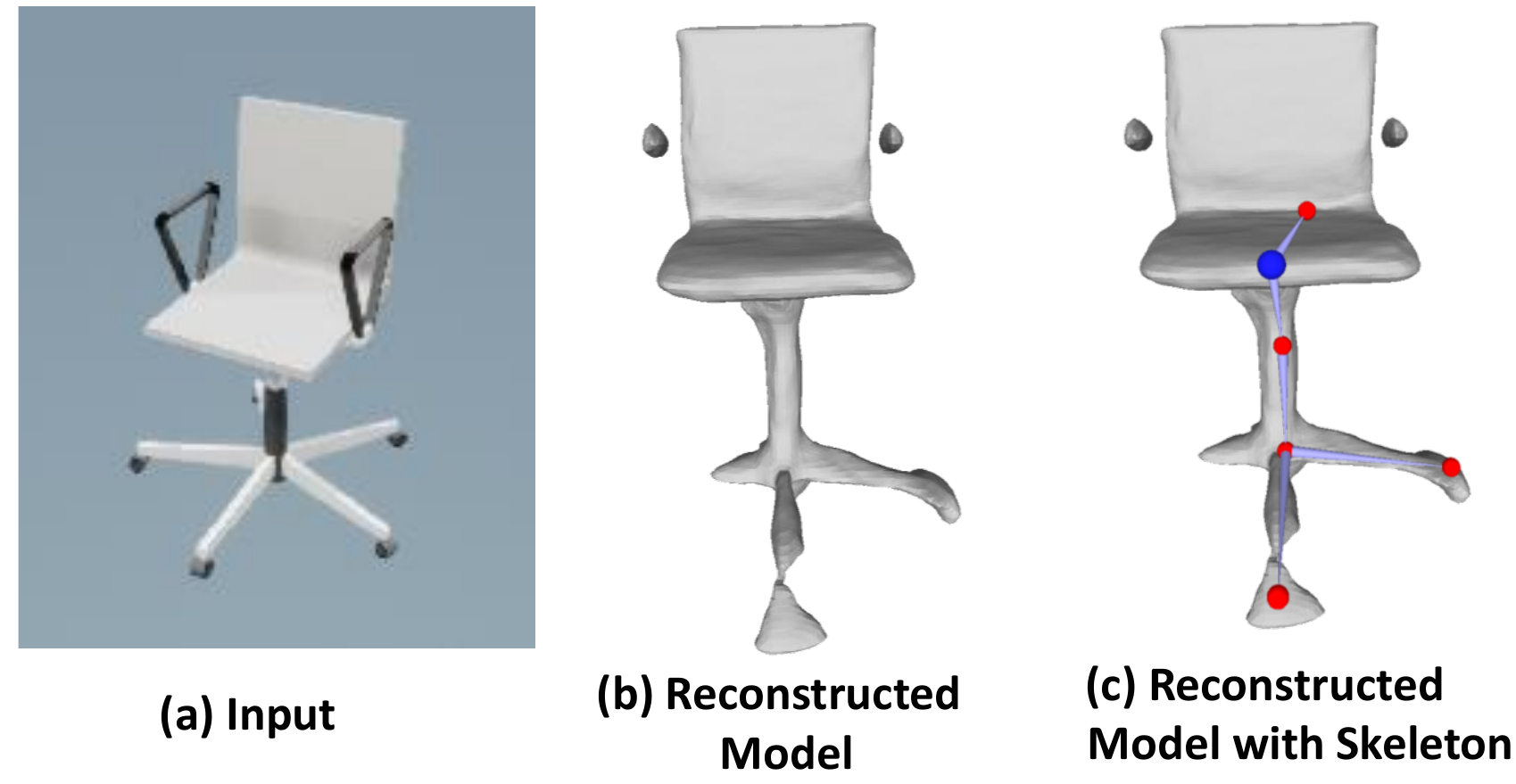}
    \end{center}
    \vspace{-6mm}
    \caption{A failure case.}
    \label{fig:failure}
    \vspace{-8mm}
\end{wrapfigure}
\vspace{-8mm}
\subsection{Failure cases.}
To achieve the goal of object wake-up and manipulate the object in the image with articulated motions, it is critical to have a well reconstructed and rigged 3D model from the input image. In Fig.~\ref{fig:failure} we show some failure cases where the quality of the rigged 3D model cannot meet the requirements for animation purposes.

\vspace{-2mm}
\section{Conclusion and Limitations}
\vspace{-2mm}
We consider an interesting task of waking up a 3D object from a single input image. An automated pipeline is proposed to reconstruct the 3D object, predict the articulated skeleton to animate the object with plausible articulations. 
Quantitative and qualitative experiments demonstrate the applicability of our work when unseen real-world images are presented at test time.

\textbf{Limitations.} First, the domain gap between synthetic to real images still exists. Second, in our current stage-wise framework, the skeleton prediction and final animation rely on the success of 3D shape reconstruction. For future work, we would like to combine shape reconstruction with skeleton embedding in a unified network structure to facilitate each task. Moreover, the collected SSkel dataset is limited in the number of objects and the range of object categories. In the future, we plan to explore its applicability in working with a much broad range of object categories and a larger number of annotated objects. 
\vspace{-2mm}
\section*{Acknowledgment}
\vspace{-2mm}
This research was partly supported by the University of Alberta Start-up Grant, the NSERC Discovery Grants, CFI-JELF grants and the Huawei-UA Joint Lab Project Grant. We also thank Priyal Belgamwar for her contribution to the dataset annotation.

\clearpage
%
\bibliographystyle{splncs04}
\bibliography{egbib}

\begin{thebibliography}{10}
\providecommand{\url}[1]{\texttt{#1}}
\providecommand{\urlprefix}{URL }
\providecommand{\doi}[1]{https://doi.org/#1}

\bibitem{amenta1999surface}
Amenta, N., Bern, M.: Surface reconstruction by voronoi filtering. Discrete \&
  Computational Geometry  \textbf{22}(4),  481--504 (1999)

\bibitem{attali1997computing}
Attali, D., Montanvert, A.: Computing and simplifying {2D} and {3D} continuous
  skeletons. Computer Vision and Image Understanding  \textbf{67}(3),  261--273
  (1997)

\bibitem{BarPop:siggraph07}
Baran, I., Popovi{\'c}, J.: Automatic rigging and animation of 3d characters.
  ACM Transactions on graphics(TOG)  \textbf{26}(3), ~72 (2007)

\bibitem{chang2015shapenet}
Chang, A.X., Funkhouser, T., Guibas, L., Hanrahan, P., Huang, Q., Li, Z.,
  Savarese, S., Savva, M., Song, S., Su, H., et~al.: Shapenet: An
  information-rich {3D} model repository. arXiv preprint arXiv:1512.03012
  (2015)

\bibitem{chen20133}
Chen, T., Zhu, Z., Shamir, A., Hu, S.M., Cohen-Or, D.: 3-sweep: Extracting
  editable objects from a single photo. ACM Transactions on Graphics (TOG)
  \textbf{32}(6),  1--10 (2013)

\bibitem{chen2019implicitfields}
Chen, Z., Zhang, H.: Learning implicit fields for generative shape modeling.
  In: IEEE/CVF Conference on Computer Vision and Pattern Recognition (2019)

\bibitem{cheriton1976finding}
Cheriton, D., Tarjan, R.E.: Finding minimum spanning trees. SIAM Journal on
  Computing  \textbf{5}(4),  724--742 (1976)

\bibitem{dosovitskiy2020image}
Dosovitskiy, A., Beyer, L., Kolesnikov, A., Weissenborn, D., Zhai, X.,
  Unterthiner, T., Dehghani, M., Minderer, M., Heigold, G., Gelly, S., et~al.:
  An image is worth 16x16 words: Transformers for image recognition at scale.
  In: International Conference on Learning Representations (ICLR) (2021)

\bibitem{fan2017point}
Fan, H., Su, H., Guibas, L.J.: A point set generation network for 3d object
  reconstruction from a single image. In: IEEE Conference on Computer Vision
  and Pattern Recognition (2017)

\bibitem{gkioxari2019meshrcnn}
Gkioxari, G., Malik, J., Johnson, J.: Mesh {R-CNN}. In: IEEE/CVF International
  Conference on Computer Vision (2019)

\bibitem{hornung2007character}
Hornung, A., Dekkers, E., Kobbelt, L.: Character animation from 2d pictures and
  3d motion data. ACM Transactions on Graphics(TOG)  \textbf{26}(1), ~1 (2007)

\bibitem{huang2013l1}
Huang, H., Wu, S., Cohen-Or, D., Gong, M., Zhang, H., Li, G., Chen, B.:
  L1-medial skeleton of point cloud. ACM Transactions on Graphics (TOG)
  \textbf{32}(4),  65--1 (2013)

\bibitem{kato2018neural}
Kato, H., Ushiku, Y., Harada, T.: Neural 3d mesh renderer. In: IEEE Conference
  on Computer Vision and Pattern Recognition (2018)

\bibitem{kearney2020rgbd}
Kearney, S., Li, W., Parsons, M., Kim, K.I., Cosker, D.: Rgbd-dog: Predicting
  canine pose from rgbd sensors. In: IEEE/CVF Conference on Computer Vision and
  Pattern Recognition. pp. 8336--8345 (2020)

\bibitem{kholgade20143d}
Kholgade, N., Simon, T., Efros, A., Sheikh, Y.: 3d object manipulation in a
  single photograph using stock {3D} models. ACM Transactions on Graphics (TOG)
   \textbf{33}(4),  1--12 (2014)

\bibitem{kulon2020weaklymesh}
Kulon, D., Guler, R.A., Kokkinos, I., Bronstein, M.M., Zafeiriou, S.:
  Weakly-supervised mesh-convolutional hand reconstruction in the wild. In:
  IEEE/CVF Conference on Computer Vision and Pattern Recognition (2020)

\bibitem{li2021d2im}
Li, M., Zhang, H.: D2im-net: Learning detail disentangled implicit fields from
  single images. In: IEEE/CVF Conference on Computer Vision and Pattern
  Recognition. pp. 10246--10255 (2021)

\bibitem{lin2018learningpoint}
Lin, C.H., Kong, C., Lucey, S.: Learning efficient point cloud generation for
  dense {3D} object reconstruction. In: AAAI Conference on Artificial
  Intelligence (2018)

\bibitem{lin2020sdfsrn}
Lin, C.H., Wang, C., Lucey, S.: Sdf-srn: Learning signed distance {3D} object
  reconstruction from static images. In: Advances in Neural Information
  Processing Systems (2020)

\bibitem{liu2019soft}
Liu, S., Li, T., Chen, W., Li, H.: Soft rasterizer: A differentiable renderer
  for image-based 3d reasoning. In: IEEE/CVF International Conference on
  Computer Vision (2019)

\bibitem{mescheder2019occupancy}
Mescheder, L., Oechsle, M., Niemeyer, M., Nowozin, S., Geiger, A.: Occupancy
  networks: Learning 3d reconstruction in function space. In: IEEE/CVF
  Conference on Computer Vision and Pattern Recognition (2019)

\bibitem{mi2020ssrnet}
Mi, Z., Luo, Y., Tao, W.: Ssrnet: scalable {3D} surface reconstruction network.
  In: IEEE/CVF Conference on Computer Vision and Pattern Recognition (2020)

\bibitem{mo2019partnet}
Mo, K., Zhu, S., Chang, A.X., Yi, L., Tripathi, S., Guibas, L.J., Su, H.:
  Partnet: A large-scale benchmark for fine-grained and hierarchical part-level
  3d object understanding. In: IEEE/CVF Conference on Computer Vision and
  Pattern Recognition (2019)

\bibitem{niemeyer2020differentiable}
Niemeyer, M., Mescheder, L., Oechsle, M., Geiger, A.: Differentiable volumetric
  rendering: Learning implicit {3D} representations without {3D} supervision.
  In: IEEE/CVF Conference on Computer Vision and Pattern Recognition (2020)

\bibitem{pan2019TMnet}
Pan, J., Han, X., Chen, W., Tang, J., Jia, K.: Deep mesh reconstruction from
  single rgb images via topology modification networks. In: IEEE International
  Conference on Computer Vision (2019)

\bibitem{park2019deepsdf}
Park, J.J., Florence, P., Straub, J., Newcombe, R., Lovegrove, S.: Deepsdf:
  Learning continuous signed distance functions for shape representation. In:
  IEEE/CVF Conference on Computer Vision and Pattern Recognition (2019)

\bibitem{peng2020convolutional}
Peng, S., Niemeyer, M., Mescheder, L., Pollefeys, M., Geiger, A.: Convolutional
  occupancy networks. In: European Conference on Computer Vision (ECCV). pp.
  523--540 (2020)

\bibitem{qi2017pointnetplus}
Qi, C.R., Yi, L., Su, H., Guibas, L.J.: Pointnet++: Deep hierarchical feature
  learning on point sets in a metric space. In: Advances in Neural Information
  Processing Systems (2017)

\bibitem{riegler2017octnet}
Riegler, G., Osman~Ulusoy, A., Geiger, A.: Octnet: Learning deep {3D}
  representations at high resolutions. In: IEEE Conference on Computer Vision
  and Pattern Recognition (2017)

\bibitem{schaefer2006image}
Schaefer, S., McPhail, T., Warren, J.: Image deformation using moving least
  squares. ACM Transactions on Graphics(TOG)  \textbf{25}(3),  533--540 (2006)

\bibitem{ShiEtAl:CVPR21}
Shi, R., Xue, Z., You, Y., Lu, C.: Skeleton merger: an unsupervised aligned
  keypoint detector. In: IEEE conference on computer vision and pattern
  recognition (2021)

\bibitem{shi20213d}
Shi, Z., Meng, Z., Xing, Y., Ma, Y., Wattenhofer, R.: 3d-retr: End-to-end
  single and multi-view {3D} reconstruction with transformers. In: The British
  Machine Vision Conference (2021)

\bibitem{tatarchenko2017octree}
Tatarchenko, M., Dosovitskiy, A., Brox, T.: Octree generating networks:
  Efficient convolutional architectures for high-resolution 3d outputs. In:
  IEEE International Conference on Computer Vision (2017)

\bibitem{pmlr-v139-touvron21a}
Touvron, H., Cord, M., Douze, M., Massa, F., Sablayrolles, A., Jegou, H.:
  Training data-efficient image transformers and distillation through
  attention. In: International Conference on Machine Learning. vol.~139, pp.
  10347--10357 (2021)

\bibitem{tretschk2020patchnets}
Tretschk, E., Tewari, A., Golyanik, V., Zollh{\"o}fer, M., Stoll, C., Theobalt,
  C.: Patchnets: Patch-based generalizable deep implicit 3d shape
  representations. In: European Conference on Computer Vision (2020)

\bibitem{tulsiani2018multi}
Tulsiani, S., Efros, A.A., Malik, J.: Multi-view consistency as supervisory
  signal for learning shape and pose prediction. In: IEEE Conference on
  Computer Vision and Pattern Recognition (2018)

\bibitem{wang2018pixel2mesh}
Wang, N., Zhang, Y., Li, Z., Fu, Y., Liu, W., Jiang, Y.G.: Pixel2mesh:
  Generating 3d mesh models from single rgb images. In: European Conference on
  Computer Vision (2018)

\bibitem{wang2020deepoct}
Wang, P.S., Liu, Y., Tong, X.: Deep octree-based cnns with output-guided skip
  connections for {3D} shape and scene completion. In: IEEE/CVF Conference on
  Computer Vision and Pattern Recognition Workshops (2020)

\bibitem{wang2018sgpn}
Wang, W., Yu, R., Huang, Q., Neumann, U.: Sgpn: Similarity group proposal
  network for 3d point cloud instance segmentation. In: Proceedings of the IEEE
  conference on computer vision and pattern recognition. pp. 2569--2578 (2018)

\bibitem{weng2019photo}
Weng, C.Y., Curless, B., Kemelmacher-Shlizerman, I.: Photo wake-up: 3d
  character animation from a single photo. In: IEEE/CVF Conference on Computer
  Vision and Pattern Recognition. pp. 5908--5917 (2019)

\bibitem{xu2019disn}
Xu, Q., Wang, W., Ceylan, D., Mech, R., Neumann, U.: Disn: Deep implicit
  surface network for high-quality single-view 3d reconstruction. In: Advances
  in Neural Information Processing Systems (2019)

\bibitem{xu2008animating}
Xu, X., Wan, L., Liu, X., Wong, T.T., Wang, L., Leung, C.S.: Animating animal
  motion from still  \textbf{27}(5), ~1--8 (2008)

\bibitem{xu2020rignet}
Xu, Z., Zhou, Y., Kalogerakis, E., Landreth, C., Singh, K.: Rignet: Neural
  rigging for articulated characters. ACM Transactions on Graphics(TOG)
  \textbf{39}(58) (2020)

\bibitem{xu2019predicting}
Xu, Z., Zhou, Y., Kalogerakis, E., Singh, K.: Predicting animation skeletons
  for 3d articulated models via volumetric nets. In: International Conference
  on 3D Vision. pp. 298--307 (2019)

\bibitem{yin2018p2p}
Yin, K., Huang, H., Cohen-Or, D., Zhang, H.: P2p-net: Bidirectional point
  displacement net for shape transform. ACM Transactions on Graphics (TOG)
  \textbf{37}(4),  1--13 (2018)

\end{thebibliography}
\end{document}